%% file: root.tex
\documentclass[letterpaper, 10 pt, conference]{ieeeconf}  

\IEEEoverridecommandlockouts                              
\let\labelindent\relax

\input{preamble.tex}

\input{glossary.tex}

\input{notation.tex}

\newcommand{\squeezeWords}{\looseness=-1}

\let\oldtwocolumn\twocolumn \renewcommand\twocolumn[1][]{%
    \oldtwocolumn[{#1}{
    \begin{center}
           \includegraphics[width=\textwidth]{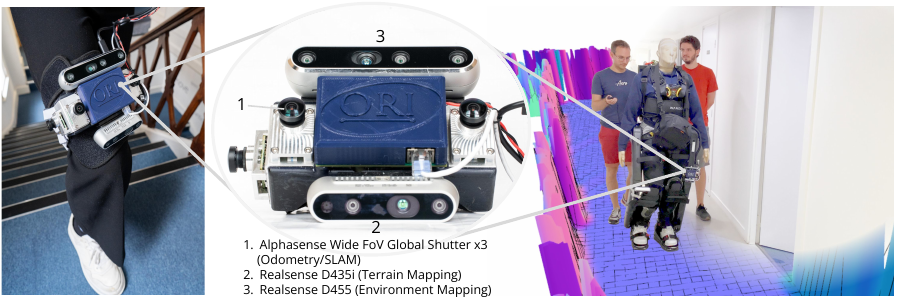}
           \captionof{figure}{\squeezeWords Exosense is a vision-based hardware and software system for scene understanding by exoskeletons. We developed a specialized multi-sensor unit (center), consisting of three global shutter wide-angle Alphasense cameras and Realsense D435i and D455 RGB-D units to provide 3D terrain and environment measurements. The hardware can be worn as a human-leg-mounted wearable device (left) or as an lower-limb attachment for an exoskeleton, such as the Wandercraft's \emph{Personal Exoskeleton} (right).}
           \label{fig:cover}
        \end{center}
    }] }

\begin{document}
\title{ \LARGE \bf Exosense: A Vision-Based Scene Understanding System For Exoskeletons}

\author{Jianeng Wang$^1$,  Matias Mattamala$^1$, Christina Kassab$^1$, Guillaume Burger$^2$, 
Fabio Elnecave$^2$, Lintong Zhang$^1$,\\ Marine Petriaux$^2$, and
Maurice Fallon$^1$ 
\thanks{\hspace{-1em}$^{1}$Oxford Robotics Institute, Dept. of Engineering Science, Uni. of Oxford, UK. \texttt{\{jianeng, matias, christina, lintong, mfallon\}@robots.ox.ac.uk } \newline%
$^{2}$Wandercraft SAS, 5 Rue Pernelle, Paris, France. \texttt{\{guillaume.burger, fabio.elnecave, marine.petriaux\}@wandercraft.health }}}

\maketitle

\begin{abstract}
  Self-balancing exoskeletons are a key enabling technology for individuals with mobility impairments. While the current challenges focus on human-compliant hardware and control, unlocking their use for daily activities requires a scene perception system. In this work, we present \emph{Exosense}, a vision-centric scene understanding system for self-balancing exoskeletons. We introduce a multi-sensor visual-inertial mapping device as well as a navigation stack for state estimation, terrain mapping, and long-term operation. We tested Exosense attached to both a human leg and Wandercraft's \emph{Personal Exoskeleton} in real-world indoor scenarios. This enabled us to test the system during typical periodic walking gaits, as well as future uses in multi-story environments. We demonstrate that Exosense can achieve an odometry drift of about \SI{4}{\centi\meter} per meter traveled, and construct terrain maps under \SI{1}{\centi\meter} average reconstruction error. It can also work in a visual localization mode in a previously mapped environment, providing a step towards long-term operation of exoskeletons. 
\end{abstract}


\section{Introduction}
Recent advances in self-balancing exoskeletons, such as Wandercraft's \emph{Atalante}
exoskeleton~\cite{Gurriet2018Wandercraft}, are enabling individuals with lower-limb
disabilities to walk independently without requiring additional support from crutches~\cite{Huynh2021Versatile}. These powered exoskeletons are being used in controlled clinical and therapeutic contexts~\cite{tian2024exo}. However, the ultimate goal is to enable users to do everyday activities at home and outdoors.

Exoskeleton development has primarily focused on hardware and control challenges, aiming to design systems that can support and transport individuals while mimicking natural human walking.
Many of these approaches employ control schemes with pre-defined gait trajectories~\cite{Yan2015}. This requires manual activation by an operator using a control panel, which increases the metabolic cost~\cite{Kim2022Visual}. 
Integrating perception systems into the loop could reduce metabolic costs by partly automating low-level control tasks, such as switching gait modes, crossing doorways, or climbing stairs.

Vision sensing, because of its rich source of information~\cite{Al-dabbagh2020}, has been the primary sensor used to achieve this. Cameras have been used to estimate the semantic class of the terrain (namely stairs, ramps, and level ground walking)~\cite{Kurbis2022,Karacan2020}, to determine basic geometric features such as ground inclination and step height~\cite{depth_vision_terrain_detection}, as well as to detect potential obstacles in the environment~\cite{Liu2021}. While these approaches are effective in providing the instantaneous information required for low-level decision making, they do not aim to integrate this information in long-term representations, which could be reused when revisiting environments.

In this work, we take initial steps towards developing long-term operation of self-balancing exoskeletons by presenting \emph{Exosense}, a vision-centric scene understanding system. Exosense aims to generate home-scale, rich scene representations from vision and geometry, capturing terrain structure, semantics, and traversability for localization and future navigation in previously visited environments. Our solution is primarily designed for self-balancing exoskeletons, such as the Wandercraft's  \emph{Personal Exoskeleton} (\figref{fig:cover}), which aims to be the first self-balancing exoskeleton designed for domestic use. To achieve this, we introduce a versatile multi-sensor unit that can be attached to the exoskeleton's leg or carried by a human as a leg-mounted wearable device. This placement allowed us to rigidly attach the Exosense to the exoskeleton, while also avoiding potential occlusions from the user's body. Additionally, it enabled us to develop and test Exosense in realistic human walking scenarios and to seamlessly transfer the system to the exoskeleton given the similarities we observed in the gait dynamics (\figref{fig:imu_measurements}). 

The key contributions of our work are:
\begin{itemize}
\item A versatile, leg-mounted multi-sensor unit that provides wide-angle vision and depth sensing for state estimation, terrain mapping, and localization.
\item A scene understanding system that builds local maps embedded with the terrain geometry, room semantics, and traversability of indoor environments.
\item A study of the performance of visual odometry for different camera configurations during typical walking patterns, where we obtained \SI{4}{\centi\meter} drift per meter traveled for the selected odometry algorithm.
\item Extensive experiments of the Exosense's scene representation, particularly accuracy of the terrain reconstruction and traversable space estimates. 
\item A real-world demonstration of the Exosense integrated into the Wandercraft's \emph{Personal Exoskeleton} for indoor localization tasks, demonstrating the potential for future long-term operation in home environments.
\end{itemize}

\section{Related Work}\label{sec:related_work}
We briefly review works on vision systems for wearable devices (\secref{sec:vision_wearable_review}) and perception for self-balancing exoskeletons (\secref{sec:perception_exoskeleton_review}), which are relevant to Exosense.

\subsection{Wearable Vision-based Systems}
\label{sec:vision_wearable_review}
Vision-based systems provide richer information about the users and their surroundings than proprioceptive sensing, drawing increasing interest in wearable robotics research~\cite{Gionfrida2024wearable}. 
Integrating computer vision into upper-limb wearable robots has been commonly used in rehabilitation applications to assist object manipulation tasks including determining object dimensions \cite{Hu2024pointgrasp} and detecting user intention \cite{Rho2024multiple}. These solutions are integrated as external egocentric cameras (e.g., mounted on glasses \cite{Kim2019eyes}) or directly attached to a robot \cite{Kuhn2019towards}. In addition to manipulation tasks, vision-based wearables have enabled independent navigation for visually-impaired people~\cite{Wang2017Enabling}. In industrial settings, similar solutions have provided assistance to alleviate joint stress and to protect users from work-related musculoskeletal disorders~\cite{Missiroli2024integrating}.

Lower-limb exoskeletons mainly use vision to detect relevant ground features for 
reliable locomotion. Ramanathan \emph{et al.}~\cite{Ramanathan2022} developed a
vision-based perception system to enable exoskeletons to change the step size subject to detected obstacles. Follow-up work extended this approach to terrain recognition~\cite{Ramanathan2024Heuristic}. Tricomi \emph{et al.}~\cite{Tricomi2023environment} used RGB images to identify types of terrain and adjust the walking controller of a hip exosuit. Karacan \emph{et al.}~\cite{Karacan2020} used a depth camera and an IMU mounted on a subject's waist to
classify objects (e.g., ramps, staircases, obstacles), and to predict staircase height. A similar setup has been deployed on a lower-limb prosthesis to recognize features in the environment \cite{Zhang2019Environmental}.

In general, the aforementioned vision-based approaches focus on assisting short-horizon low-level decision making, without preserving historical data. With Exosense we instead aim to build a map representation that can be reused for future operations in indoor settings.

\subsection{Perception for Self-balancing Exoskeletons}
\label{sec:perception_exoskeleton_review}

Existing self-balancing exoskeletons primarily rely on proprioceptive sensing for state estimation. Vigne \emph{et al.}~\cite{Vigne2020state} incorporated multiple IMUs and robot joint encoder measurements to estimate position and velocity through a flexible kinematic model to account for deformations during exoskeleton walking. MOVIE~\cite{Vigne2022movie} fused the same sensor measurements and takes a velocity-aided approach to estimate the robot's orientation with respect to gravity. Elnecave \emph{et al.}~\cite{Xavier2023multi} built on top of MOVIE to estimate the 6 DoF pose and velocity of the exoskeleton's body using an EKF. 

Self-balancing exoskeletons also incorporate proprioceptive sensing into the control scheme. Tian \emph{et al.} \cite{Tian2024dual} used motor force sensors to estimate the center of mass deformation and physical parameters of the human operator during exoskeleton walking, which were then fed into a joint control framework to help the robot walk stably. Li \emph{et al.} \cite{Li2024human} used IMUs to estimate both robot and human center of mass; these estimates were integrated into a human-in-the-loop cooperative control scheme to adaptively adjust the motion controller to follow the user's intention.

In contrast to previous work that is tightly tailored to the exoskeleton platform, we developed an integrated sensing unit that relies on exteroceptive sensing only, hence being independent of the particular platform or user gait.
\begin{figure}
  \centering
  \includegraphics[width=0.9\linewidth]{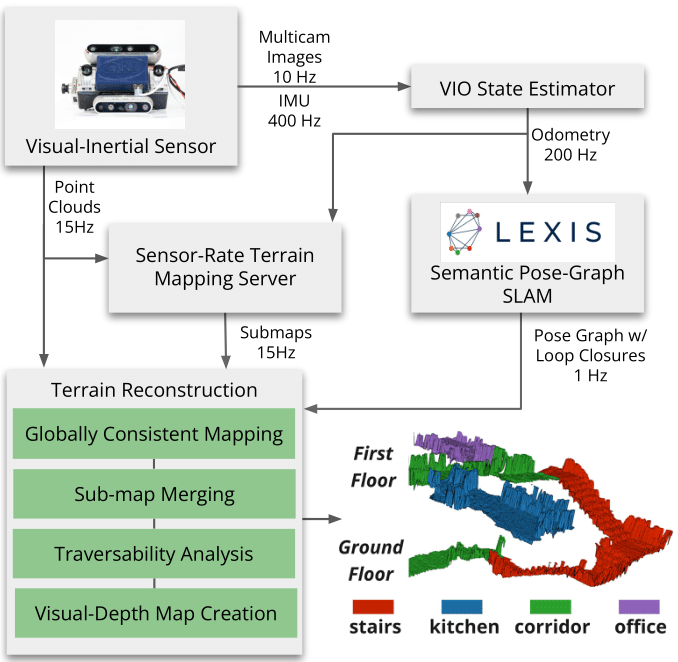}
  \caption{Exosense scene understanding system. The inputs are RGB images and 3D point clouds from the multi-sensor unit. Different modules provide odometry and sensor-rate terrain maps, which are integrated into a semantic pose graph and processed with a terrain reconstruction module. The final scene representation (bottom right) contains terrain geometry, semantics, and traversability as well as visual localization information to aid long-term operation.}
  \label{fig:pipeline}
  \vspace{-10pt}
\end{figure}

\section{System}\label{sec:system}
The Exosense system overview is shown in \figref{fig:pipeline}. It is a scene understanding system which involves a leg-mounted vision-based sensing unit, and a navigation stack designed for highly dynamic walking motions. The system generates an environment representation encoding the terrain geometry, semantics, and traversability that can enable the continuous deployment of the exoskeleton for localization and navigation. The following sections describe the main components, from the hardware design to the navigation stack.

\subsection{Multi-sensor Setup}\label{subsec:hardware}
The Exosense hardware consists of a lower-limb-mounted wearable device, shown in \figref{fig:cover}. The sensing device includes a Sevensense Alphasense unit with three hardware-synchronized wide-angle global shutter cameras plus an inertial sensor, and two Realsense RGB-D cameras (D435i and D455). The Alphasense unit is used mainly for state estimation (odometry estimation and localization) due to its wide field-of-view (FoV). The Realsense cameras provide 3D sensing for terrain mapping.

The multi-sensor unit can be attached to either a human thigh or an exoskeleton. We developed the sensing system to only require exteroceptive sensing so as to disregard challenges related to leg deformation and bending. On the exoskeleton, the device is rigidly attached to the thigh to avoid occlusions with the rest of the user's body and to facilitate its future integration with the exoskeleton's walking controller.

\begin{figure}
  \centering
  \includegraphics[width=\linewidth]{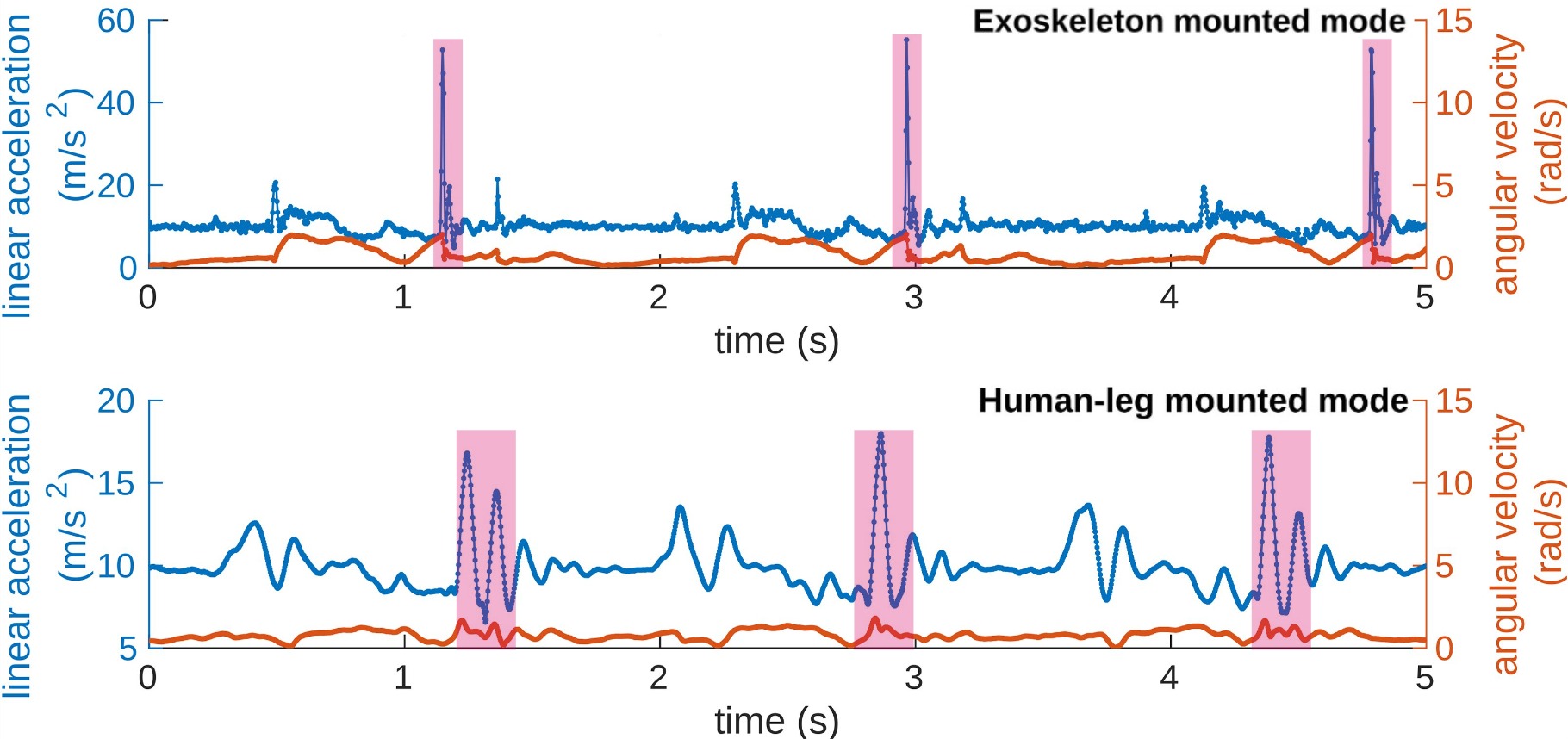}
  \caption{Sample of linear acceleration and angular rotation rates measured by Exosense in exoskeleton (top) and human-leg-mounted (bottom) configurations. Both modes have a similar gait duration. The highlighted spikes (pink) occur during foot strikes.}
  \label{fig:imu_measurements}
\end{figure}

Because the device can also be attached to a human leg, we could develop the scene understanding system without needing permanent access to an exoskeleton---enabling us to test algorithms that go beyond the current capabilities of the exoskeleton, such as multi-floor navigation. This design decision is supported by the similarities we observed in the walking motions of the human-leg-mounted and exoskeleton-mounted sequences, as illustrated in \figref{fig:imu_measurements}.

\subsection{Visual-Inertial Odometry}
To estimate ego-motion, we use a visual-inertial odometry system to provide high-frequency state estimation and to handle high rotation rates and jerk during the exoskeleton locomotion. We considered OpenVINS \cite{Geneva2020OpenVINS}, VILENS-MC
\cite{Zhang2022VILENS-MC} and ORB-SLAM \cite{Campos2021ORB-SLAM}. We evaluated their performance in custom
sequences recorded with our multi-sensor unit to assess their performance and reliability under walking patterns, which is discussed in
\secref{sec:experiment-odometry}. For our experiments, we chose OpenVINS due to its better balance of estimation accuracy and computational cost.

\subsection{Semantic Pose-graph SLAM}
The odometry estimate serves as input to a visual SLAM system. We used LEXIS~\cite{Kassab2024Lexis}, as it provides a semantic pose graph representation that can be easily extended with other information sources, such as terrain maps.
LEXIS constructs a pose graph representation with evenly spaced nodes. The nodes store corresponding RGB images, which are used for visual localization and loop closure detection. For semantics, LEXIS uses the CLIP visual-language model~\cite{Radford2021} to obtain visual embeddings, which are compared against text embeddings of a list of room classes (e.g., office, kitchen, corridor), providing potential room labels for each node in the graph.

The room labels associated to each node enable hierarchical place recognition by comparing the predicted room class of the current image to the keyframe labels in the graph. PnP~\cite{Fischler1981} is used as a geometric verification step to propose loop closure candidates, which are jointly optimized in a factor graph to reduce the drift in the graph.

The output of LEXIS is a pose graph encoding odometry and loop closure connectivity with a pose-level room segmentation. In Exosense we extend this representation by adding terrain maps to each node, which can be refined by exploiting the room information. This enables us to produce an \emph{elastic} globally-consistent terrain representation defined by the pose graph, as proposed in the Atlas framework~\cite{Bosse2004Atlas}.

\subsection{Mapping and Reconstruction}
\label{subsec:mapping-reconstruction}
To obtain the local terrain maps, we use the method by Fankhauser \emph{et al.}~\cite{Fankhauser2018Elevation}, which integrates point clouds from both RGB-D cameras at \SI{15}{\hertz}, to generate a rolling local multi-layered 2.5D map~\cite{Fankhauser2016GridMap} at \SI{2}{cm} resolution around
the Exosense's sensing unit location. We used this method as a server, providing local terrain maps to be attached to LEXIS' pose graph nodes on request.

While this representation enables a lightweight elastic terrain reconstruction, the individual
submaps only represent a local region around the corresponding node, which might
be suboptimal due to moving objects and partial visibility. Hence, we propose to exploit the semantic room information already stored in the pose graph to refine the terrain estimates, creating single \emph{room-based} terrain maps.

For submaps within the same room, we 
fuse the height values of overlapping map cells using the median:
\begin{equation}\label{eq:merge}
h_{i}^{\text{merged}} = \text{median}(\{h_i\}),
\end{equation}
where $h_{i}^{\text{merged}} $ is the fused height value for the $i^{\text{th}}$
cell of terrain map, $\{h_i\}$ is the set of all the valid overlapping height
values in cell $i$. While other submap fusion strategies could be used, here we exploit the room information already provided by LEXIS, which provides a semantic guidance. 

\subsection{Terrain Traversability Analysis}
\label{sec:terrain_analysis}
Following the room-based fusion step, we wish to estimate the traversability of each room's terrain map. This traversability estimate is computed on a cell basis, characterizing which areas of each room should be accessible by the exoskeleton in a navigation setting. To obtain it, we perform a geometric analysis of the local terrain tailored to the exoskeleton's gait specifications.

Technically, the local terrain analysis module determines a traversability score for a cell
$i$, $t_{i}\in [0,1]$, which characterizes how difficult it would be for the exoskeleton
to step on the cell (specifically, $0$ for untraversable and $1$ for traversable). 
For this, we assume that the exoskeleton has a nominal maximum stride length $s^{*}$ (size of step forward) and step height $h^{*}$ (maximum height it can step on). For each cell $i$ with height $h_{i}$, we select the neighboring cells $j$ within a radius $s^{*}$,
denoted by $\mathcal{C}_{s^{*}}$, and compute the maximum height difference $h_{i}^{\text{max}}$ in the neighborhood:
\begin{equation}
  \label{eq:step_height}
  h_{i}^{\text{max}} =\text{max}( |h_{j} - h_{i}|), j \in \mathcal{C}_{s^{*}}.
\end{equation}
We then define the traversability score of a cell as the percentage difference of the maximum height difference $h_{i}^{\text{max}}$ with respect to the nominal step height $h^{*}$:
\begin{equation}\label{eq:traversability}
t_{i} = 1 - \text{min}\left( \frac{h_{i}^{\text{max}}}{h^{*}}, 1 \right).
\end{equation}
This traversability score then represents a conservative cell estimate of how safe it would be for the exoskeleton to step into any other cell given this nominal step height. \figref{fig:traversability_compare} illustrates how this compares to a geometric approach based on surface normals on a staircase.

\subsection{Localization within the Scene Representation}
Exosense generates a scene representation that includes terrain geometry, semantics, and traversability. We extend it with images and depth maps obtained at each topological map node. This enables us to perform place recognition and metric localization in subsequent missions, enabling the reuse of previously built maps.

Our localization approach uses visual bags of words~\cite{dbow} to obtain place candidates, and PnP~\cite{Fischler1981} to obtain a metric pose estimate from the RGB and depth images. We combine this with the odometry estimate to provide a continuous pose estimate between relocalizations, as well as providing an estimate when the exoskeleton visits an unmapped area. We demonstrate this in \secref{sec:demo}.

\def\thesubsectiondis{Exp \Alph{subsection}.} 
\renewcommand{\thesubsection}{\thesection-Exp \Alph{subsection}}

\section{Experiments}
\label{sec:experiment}

We conducted several experiments to validate the Exosense's multi-sensor unit and navigation pipeline for indoor exoskeleton applications. We collected two datasets using the Exosense multi-sensor unit in two indoor environments: 
\begin{itemize}[leftmargin=*]
\item \emph{Human} -- Four sequences recorded with the Exosense device mounted on a human leg (\figref{fig:cover}, left): Sequences with 2-minute duration \emph{H1, H2, H3} were captured using a Vicon motion capture system in a research lab, while \emph{H4} was 10-minute long and recorded in a multi-floor office environment. The objective was to evaluate our design before testing on the exoskeleton, as well as testing \textit{Exosense} mapping capabilities which go beyond the current exoskeleton locomotion capabilities.
\item \emph{Exo} -- Two 7-minute sequences recorded with the Exosense device attached to the thigh of Wandercraft's \emph{Personal Exoskeleton} (\figref{fig:cover}, right): \emph{E1} was used to evaluate the terrain reconstruction, while \emph{E2} was used to demonstrate the localization capabilities of Exosense. The exoskeleton was teleoperated while carrying a dummy during the recordings in a mixed office and lab environment with occasional passers-by. The objective of this dataset was to assess the mapped terrain quality in realistic conditions, as well as the potential of Exosense for indoor navigation.
\end{itemize}

All data was post-processed with a mid-range laptop, Intel i7 10750H @ 2.60Hz 12 core laptop, Nvidia GTX 1650Ti GPU.
All the algorithms are CPU-based except for the CLIP feature extractor in LEXIS.

Our first two experiments, \secref{subsec:ex_hardware} and \secref{sec:experiment-odometry}, aim to assess our hardware and odometry estimators decisions prior to the deployment of Exosense on the exoskeleton---hence they are demonstrated with the human-leg mounted mode.
We additionally show the potential of the full pipeline to operate in multi-story environments in \secref{subsec:multistory_mapping}. The last three experiments assess the reconstruction accuracy, traversability quality, and demonstrate a localization use case in the \emph{Exo} sequences. 

\subsection{Study of Wide FoV Multi-Camera Systems \textit{[Human]}}
\label{subsec:ex_hardware}
Our first experiment tested the suitability of our multi-camera setup. We achieved this by comparing the performance of a visual-inertial odometry system, for different numbers of cameras and different FoV in sequence \emph{H1}. We used VILENS-MC~\cite{Zhang2022VILENS-MC}, a fixed-lag optimization-based visual-inertial odometry algorithm designed in our group which works with multi-camera systems. The Exosense sensing unit was carried by a human as a wearable device on the person's thigh, while mimicking the exoskeleton walking pattern. Both wide and narrow field-of-view (FoV) images were recorded, and a Vicon motion capture system was used to provide ground truth poses. When evaluating the odometry performance, we report the \gls{rmse} of the \gls{rpe} as our main metric.

\input{table_FOV}

\tabref{tb:stat_RPE_FOV} reports the main results of this experiment. We observe that using a small FoV camera results in significant drift or even motion tracking failure due to the high accelerations and jerks present in the walking motion. Wider FoV cameras help to mitigate these effects, significantly reducing drift. Adding the lateral camera mitigates situations where no features are detected in front of the device. The lowest drift rates are achieved using both wide FoV cameras and the multi-camera setup with forward and lateral views. This configuration achieved reliable motion tracking, even in scenarios with significant viewpoint changes or occlusions under the jerky walking motion. 

\subsection{Comparison of VI Odometry Algorithms [Human]}
\label{sec:experiment-odometry}

Next, we extended the evaluation of odometry estimators for Exosense to other open source algorithms, using the wide FoV and 3-camera configuration determined in the previous experiment. The objective was to assess the performance of other methods in these challenging walking conditions. We compared VILENS-MC to ORB-SLAM (optimization-based)~\cite{Campos2021ORB-SLAM} and OpenVINS (filtering-based)~\cite{Geneva2020OpenVINS}. We must note that for ORB-SLAM we used the stereo-inertial configuration as it does not support multi-camera setups; we also disabled loop closure mechanisms for a fair comparison with the odometry systems. Further, OpenVINS processes each camera as a monocular input, while VILENS-MC treats the three cameras as a stereo pair and a monocular camera.

We used sequences \emph{H2} and \emph{H3} to test the performance of these systems in new conditions not considered in \emph{H1}. In \textit{H2}, the operator walked at a slow pace in a loop that included a small staircase. The sequence had a peak linear acceleration of \SI{37.6}{\meter\per\square\second} and rotation rate of \SI{4.4}{\radian\per\second}. In \textit{H3} we included periods of abrupt rotation change and occasional occlusions of the front stereo cameras. This sequence had peak acceleration and rotation rates of \SI{55.1}{\meter\per\square\second} and \SI{5.0}{\radian\per\second} respectively. The algorithms were run five times on each sequence. \gls{rpe} at \SI{1}{\meter} and \SI{5}{\meter} are presented in~\tabref{tb:stat_RPE_1m5m}. 

\begin{figure}
  \centering
  \includegraphics[width=\linewidth]{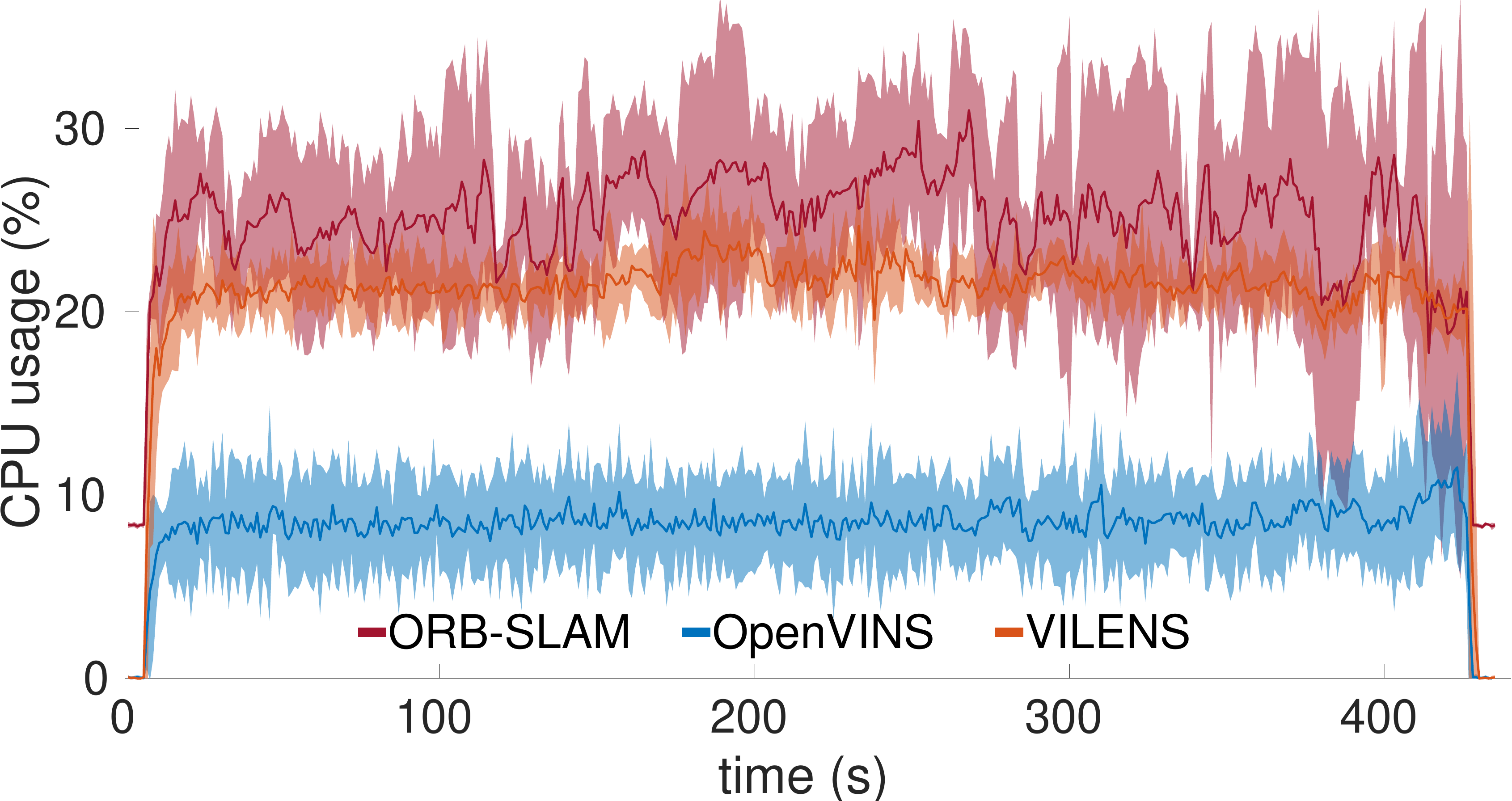}
  \caption{Exp B -- CPU usage over time for the evaluated odometry algorithms over five runs. The darker lines show the mean, while the shaded areas are the 95\% confidence interval.
 OpenVINS is significantly more lightweight---using about half the computation.}

  \label{fig:cpu_usage}
\end{figure}

Additionally, to provide further insights on the computational budget required by each method, we logged the CPU usage of the algorithms. These results are presented in \figref{fig:cpu_usage}.

\input{table_odom}

Our odometry evaluation results show that the three tested VIO algorithms are robust and reliable even during walking patterns. This follows our design decision to use wide FoV cameras as discussed in \secref{subsec:ex_hardware}. ORB-SLAM experiences higher drift rates at times as it lacks multi-camera support, which we also noted in a previous paper \cite{Zhang2022VILENS-MC}. While OpenVINS and VILENS-MC achieve comparable odometry accuracy (\SI{20}{\cm} RPE at \SI{5}{\meter}, i.e. \SI{4}{\cm} drift per meter traveled), OpenVINS uses less computation (i.e., under \SI{10}{\%} CPU usage). As a result, we chose OpenVINS as the odometry source for Exosense for our next experiments. In future we envisage these algorithms being run on low power hardware such as an ASIC or FPGA chip.

\subsection{Multi-story Mapping with Exosense \textit{[Human]}}
\label{subsec:multistory_mapping}
The last experiment in the human-leg mounted mode tested the full Exosense navigation pipeline for sequence \emph{H4}, featuring a multi-story building. This sequence shows the potential of Exosense to build multi-floor representations that are beyond the current locomotion capabilities of self-balancing exoskeletons on multi-step staircases.

\begin{figure}
  \centering
  \includegraphics[width=\linewidth]{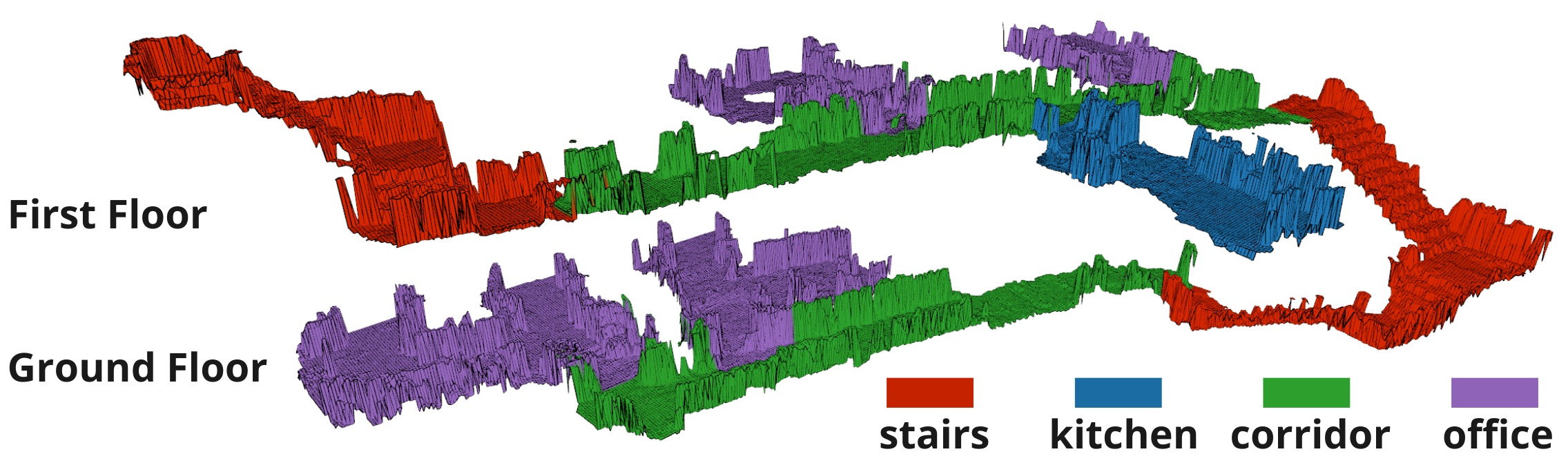}
  \caption{Exp C -- Multi-story mapping of sequence \emph{H4} in the \emph{Home} dataset. Exosense generated a globally consistent multi-floor terrain map. Each room is a single individual submap colored by its type.}
  \label{fig:mapping_semantics}
\end{figure}

\figref{fig:mapping_semantics} demonstrates the terrain reconstruction of a multi-story office environment carrying the Exosense unit on a human leg. The system handles elevation maps in the multi-floor scenario and merges them by semantic room labels, which yields a smooth terrain map. We demonstrate the detailed view of multi-floor mapping in the attached multi-media material and present the quantitative reconstruction evaluation in \secref{sec:experiment-mapping}.

\setlength{\belowcaptionskip}{-4pt}

\subsection{Evaluation of Terrain Reconstruction Quality \textit{[Both]}}
\label{sec:experiment-mapping}
We evaluated the Exosense terrain reconstruction quality in both the exoskeleton and human-leg mounted modes. For this experiment we focused on staircases present in sequences \emph{H4} (\emph{Human}) and \emph{E1} (\emph{Exo}), using millimeter-accurate reconstructions from tripod-based laser scanners, Leica RTC360 and Faro Focus 3D-X130, respectively. We ran Exosense with the same setup used in \secref{subsec:multistory_mapping}.

\begin{figure}
  \centering
  \includegraphics[width=\linewidth]{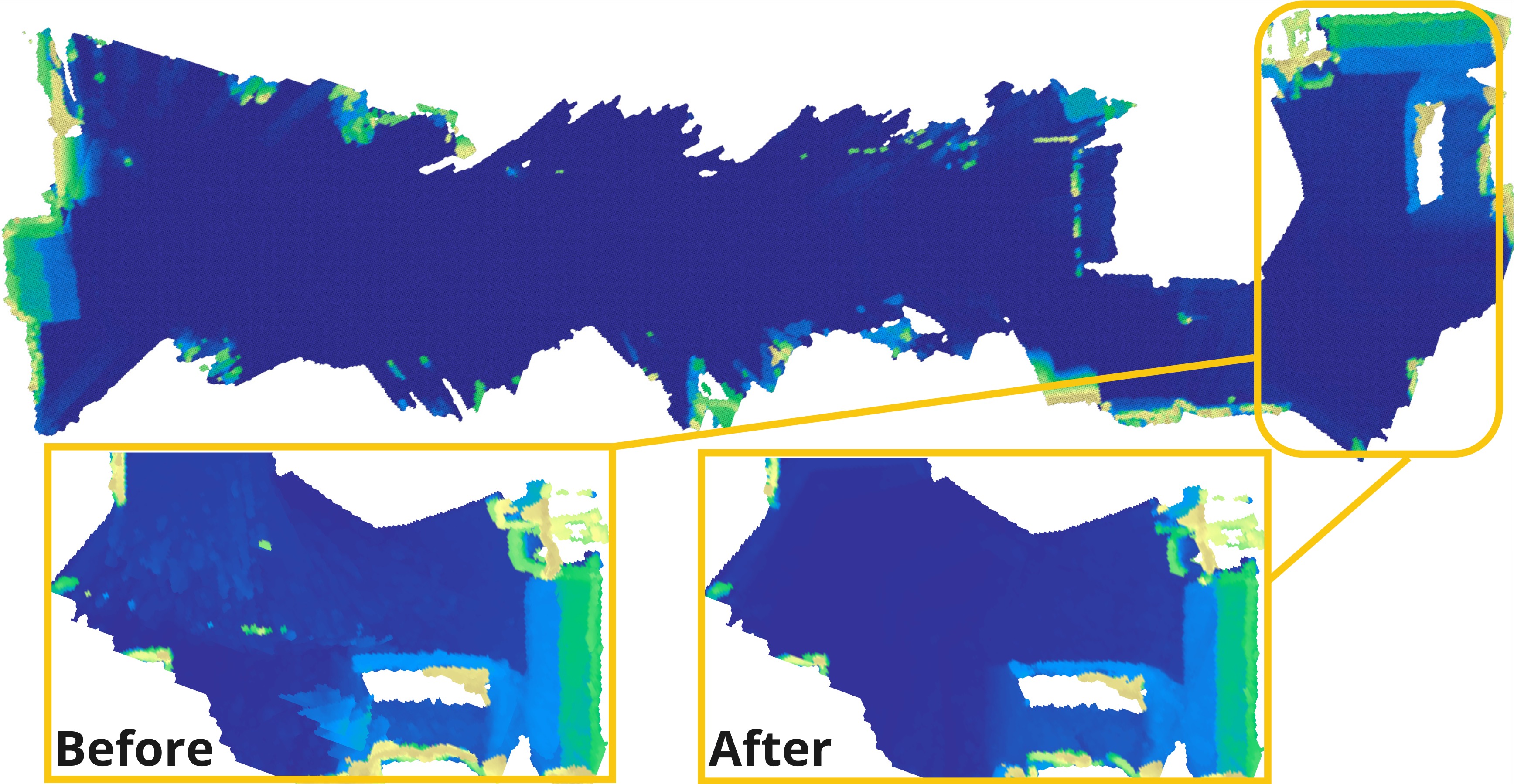}
  \caption{Exp D -- Qualitative mapping result after submap merging of the
  \emph{Exo} sequence, colored by the elevation. Staircases and part of the ground areas are shown in detail both before and after applying submap merging (bottom). The median-based merging method removed outliers in the terrain submap while preserving the sharp features and edges of terrain geometry.}
  \label{fig:mapping_demo_wandercraft}
  \vspace{-10pt}
\end{figure}

\figref{fig:mapping_demo_wandercraft} shows the reconstruction of the \emph{Exo} sequence colored by elevation. We also show maps before and after applying the submap fusion strategy introduced in \secref{subsec:mapping-reconstruction}. We observed improvements in the terrain flatness, and better crispness at the edges of the staircase steps. Our fusion strategy mitigated outliers present in the individual submaps, resulting in a consistent reconstruction of the terrain.

Further, we performed a quantitative evaluation against the laser scans, by extracting key areas from the elevation map that the exoskeleton could traverse (e.g., staircases). We cropped these regions and converted the terrain maps into meshes to preserve the geometry of the terrain, and then sampled 10000 points per square meter from meshes to compute point-to-point distances to the ground truth scans. 

\begin{figure}
  \centering
  \includegraphics[width=\linewidth]{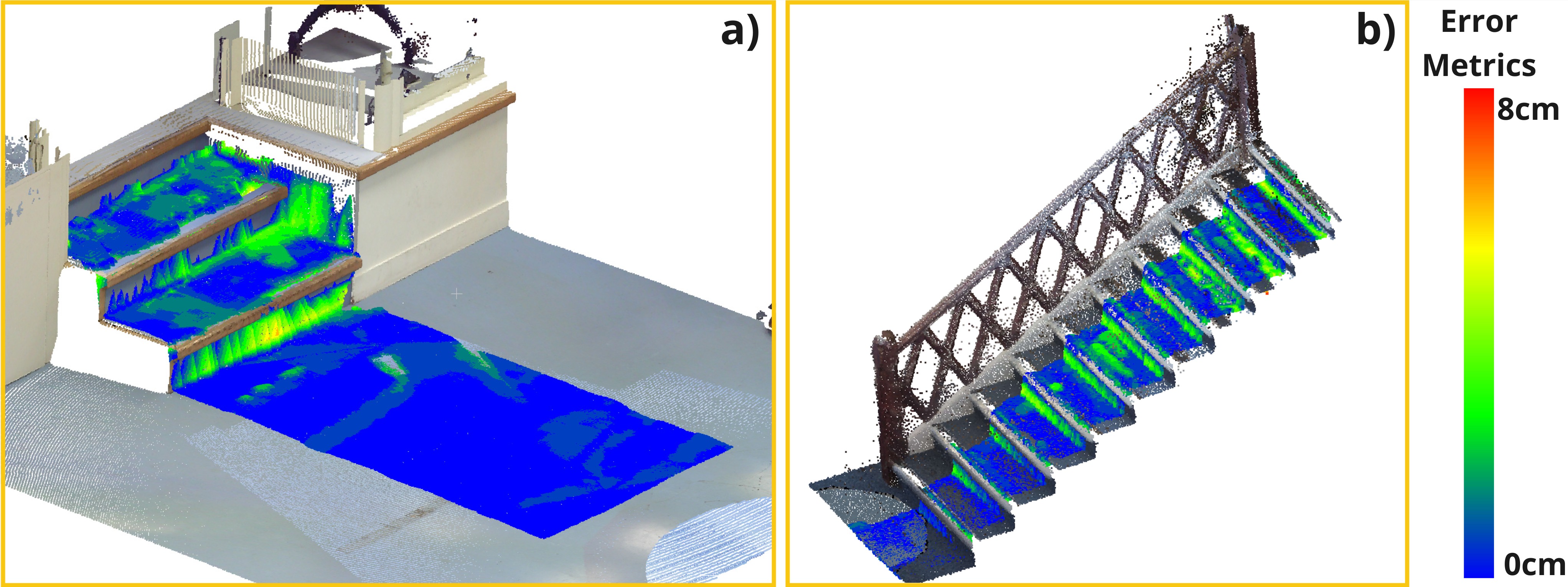}
  \caption{Exp D -- Maps of two areas of interest from the Exosense system in (\textbf{a}) exoskeleton mounted  and  (\textbf{b}) human-leg mounted modes. The mapping results are compared with ground truth laser scans and colored by the point-to-point distance. Errors in the near-vertical surfaces of the terrain map should be ignored.}
  \label{fig:mapping_detail}
  \vspace{-10pt}
\end{figure}

We present the results in \tabref{tb:accuracy}. The mapping results under both mounting modes showed similar accuracy, indicating the design choices made using the human-leg mounted mode can be successfully transferred to the exoskeleton-mounted mode. For the exoskeleton-mounted mode, Exosense produced an average error under \SI{1}{\cm} with $90^{\text{th}}$ percentile error under \SI{2}{\centi\meter}, indicating a sensible reconstruction quality for the use with the exoskeleton. \figref{fig:mapping_detail} additionally shows the error distribution for staircases present in the sequences. The large errors mainly appear on the vertical areas, which is expected from an elevation-based terrain reconstruction method and the chosen evaluation procedure.

\input{table_accuracy}

\subsection{Evaluation of Terrain Traversability \textit{[Exo]}}
\label{sec:experiment-traversability}
\begin{figure}[tbp]
  \centering
  \includegraphics[width=\columnwidth]{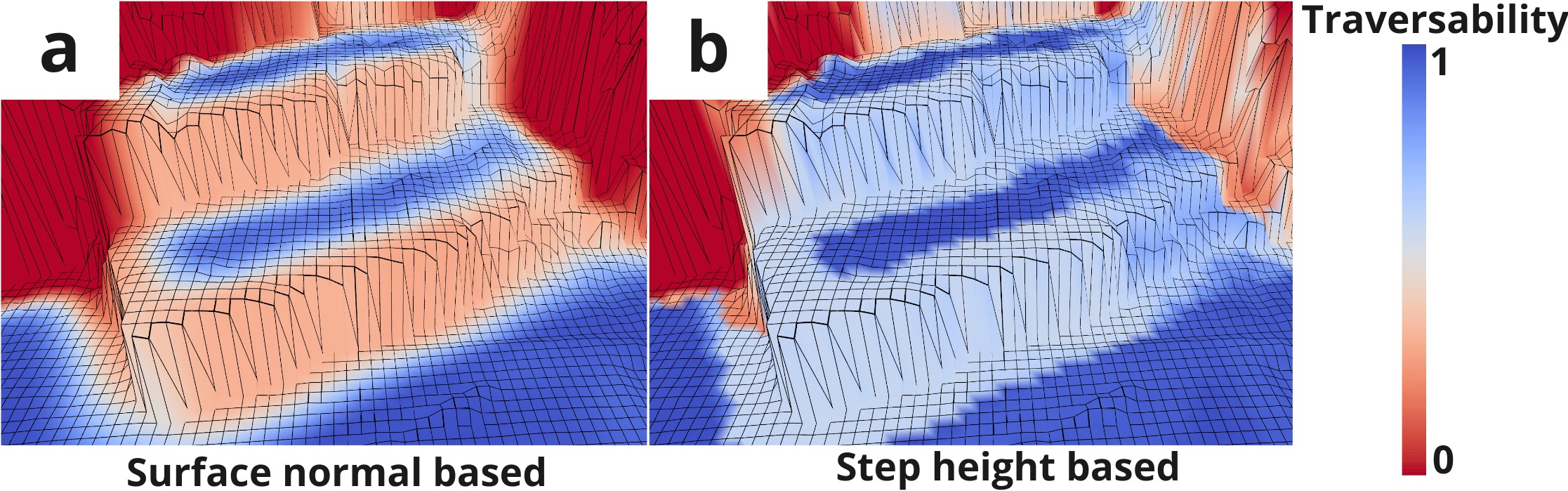}
  \caption{Exp E -- Comparison between traversability analysis obtained based on (\textbf{a})
  terrain normals, and (\textbf{b}) our method based on the exoskeleton's step height.
  We observe that for the same traversability range, our method assigns a higher traversability score to the complete staircase compared to the normals-based method.
  }
  \label{fig:traversability_compare}
  \vspace{-10pt}
\end{figure}

Next, we assessed the traversability estimation result of the Exosense system.
In Fig. \ref{fig:traversability_compare}, we present an example of per-cell traversability obtained from the mapping result in sequence \emph{E1}: On the left, we show a baseline based on surface normals~\cite{traversability2016}, while the right is our proposed step height-based method. We set our method with a maximum stride length of \SI{20}{\cm} and nominal step height of \SI{20}{\cm}. We observed that it correctly assigned high traversability scores to the riser and treads of the staircase, which reflects the effective traversable areas of a walking system. In contrast, methods based on surface normals correctly determine walls to be untraversable, but may assign low traversability scores to the risers---which is undesired for navigation tasks.

To quantitatively show the benefits of our approach, we stored the per-cell traversability predictions and estimated elevation, and hand-labeled the traversable and untraversable areas. We then binarized the traversability scores of the normals-based method and ours with different thresholds. Then we evaluated the quality of the predictions as a classification problem, where the traversable areas were positive and the untraversable areas negative. We computed precision, recall, and F-score values over different thresholds to evaluate the traversability estimation accuracy. 
Our method obtained F-scores above $0.9$ for different threshold values, with an optimal threshold of $0.5$ (F-score value $0.93$). In contrast, the normals-based method reported lower F-scores (below $0.87$) for all threshold values and was more sensitive to changes in the optimal traversability threshold.

\subsection{Localization Demonstration [Exo]}
\label{sec:demo}
Finally, we demonstrated the ability of the Exosense to visually relocalize within a prior map. To achieve this, we recorded an initial sequence using the leg mounted configuration, and then used the localization mode in the Exo-mounted configuration (sequence \emph{E2}).

\begin{figure*}[h]
  \centering
  \includegraphics[width=0.9\textwidth]{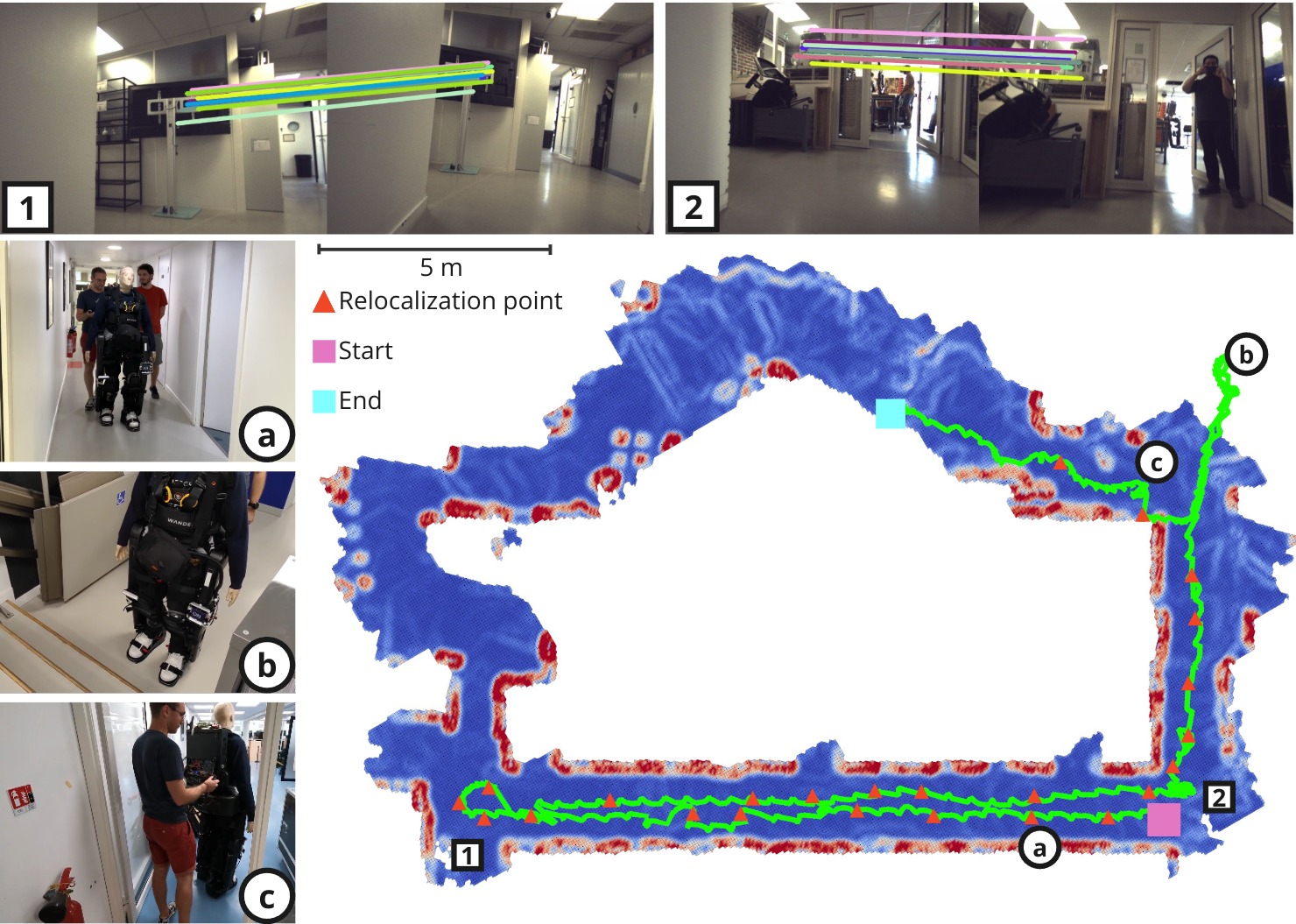}
  \caption{Exp F -- Exoskeleton-mounted localization demonstration within a prior map. 
  The green line indicates the path estimated by \textit{Exosense} using our navigation system in localization mode; the triangles denote a subset of areas where a relocalization fix was achieved. Left images show part of the testing area: \protect\boldcircled{a} while crossing a long corridor, \protect\boldcircled{b} in an unmapped region, and \protect\boldcircled{c} passing through a narrow doorway. The top images show visual matches between the prior map (human-leg-mounted, left), and live exoskeleton stream (right).}
  \label{fig:localization_demo}
\end{figure*}

\figref{fig:localization_demo} shows the prior map built with Exosense in human-leg-mounted mode. The path followed by the exoskeleton in a subsequent experiment is shown in green. The exoskeleton was teleoperated to walk around \SI{80}{\meter}, with our system obtaining about $60$ visual relocalizations (about one every \SI{0.8}{m} on average, shown as orange triangles). Our localization system was able to correct the estimate when the exoskeleton returns back along the corridor (\boldcircled{a}) and when passing through the narrow doorway (\boldcircled{c}). We observed that the odometry estimate was generally reliable when moving through unmapped regions (\boldcircled{b}), and the system was able to relocalize when returning to previously visited areas.

\def\thesubsectiondis{\Alph{subsection}.} 
\renewcommand{\thesubsection}{\thesection-\Alph{subsection}}

\section{Conclusions}\label{sec:conclusion}
We introduced Exosense, a vision-based scene understanding system for self-balancing exoskeletons. Our system consists of a multi-sensor unit and a navigation stack, designed to be independent of the exoskeleton hardware and also usable as a wearable device. We investigated the hardware and dynamics of the problem, concluding that a visual-inertial unit with wide-angle cameras overcomes most of the challenges of the walking motion. We further introduced a mapping pipeline able to capture accurate terrain structure, semantics and traversability, as well as demonstrated how Exosense can relocalize in previously visited places. This provides input on the advantages of exteroceptive sensing for the eventual deployment of exoskeletons in indoor environments. In future work, we aim to extend the applicability of Exosense for long-term, multi-session localization and mapping applications under environment changes, explore its usage in outdoor scenarios, and further integrate the system into the exoskeleton's navigation and control stack additionally exploiting its proprioceptive sensing.

\section*{ACKNOWLEDGMENT}
This work is supported by a Royal Society University Research Fellowship (Fallon, Kassab),  Horizon Europe project DigiForest 101070405 (Wang), and EPSRC C2C Grant EP/Z531212/1 (Mattamala). We thank Wayne Tubby and Matthew Graham for hardware design support.

\balance
\bibliographystyle{IEEEtran}
\bibliography{paper}

\end{document}

%% file: preamble.tex
\usepackage{times}
\usepackage{amsmath,amsfonts}
\usepackage{algorithmic}
\usepackage{algorithm}
\usepackage{array}
\usepackage{balance}
\usepackage[dvipsnames,table]{xcolor}
\usepackage{textcomp}
\usepackage{stfloats}
\usepackage{url}
\usepackage{verbatim}
\usepackage{graphicx}
\usepackage{todonotes}
\usepackage{booktabs} 
\usepackage{bbm} 
\usepackage{mathtools}
\usepackage{pgf} 
\usepackage{etoolbox} 
\usepackage{amssymb}

\usepackage{soul}

\usepackage{paralist}
\usepackage{enumitem} 

\usepackage{multicol}
\usepackage{hyperref} 
\usepackage{lipsum} 
\usepackage{siunitx} 
\usepackage{mathtools}
\usepackage{xspace}

\usepackage{multirow}
\usepackage{pbox}

\usepackage{threeparttable}
\usepackage{tablefootnote}

\usepackage{tikz}

\usepackage[normalem]{ulem}
\usepackage{gensymb}

\definecolor{MK_Two_One}{RGB}{178,24,43} 
\definecolor{MK_Two_Two}{RGB}{239,138,98}
\definecolor{MK_Two_Three}{RGB}{253,219,199}
\definecolor{MK_Two_Four}{RGB}{209,229,240}
\definecolor{MK_Two_Five}{RGB}{103,169,207}
\definecolor{MK_Two_Six}{RGB}{33,102,172} 

\hypersetup{
colorlinks=true
,linkcolor=black
,citecolor=black
,filecolor=MK_Two_Six
,urlcolor= MK_Two_Six
,menucolor=MK_Two_Five
,runcolor=MK_Two_Four
,linkbordercolor=MK_Two_One
,citebordercolor=MK_Two_Two
,filebordercolor=MK_Two_Three
,urlbordercolor=MK_Two_Six
,menubordercolor=MK_Two_Five
,runbordercolor=MK_Two_Four
}

\usepackage{subcaption}
\definecolor{high}{RGB}{116, 173, 209}  
\definecolor{low}{RGB}{244, 109, 67}  

\def\secref#1{Sec.~\ref{#1}}

\def\figref#1{Fig.~\ref{#1}}
\def\tabref#1{Tab.~\ref{#1}}
\def\eqref#1{Eq.~(\ref{#1})}

\usepackage{tikz}
\newcommand*\notecircle[1]{\tikz[baseline=(char.base)]{
		\node[shape=circle,draw,inner sep=1pt, thick] (char) 
		{\footnotesize{#1}};}}
\newcommand{\boldcircled}[1]{\notecircle{\textbf{#1}}}

%% file: glossary.tex
\usepackage[abbreviations]{glossaries-extra}

\glssetcategoryattribute{abbreviation}{indexonlyfirst}{true}

\glssetcategoryattribute{abbreviation}{nohyperfirst}{true}

\newabbreviation{auroc}{AUROC}{Area Under the Receiver Operating Characteristic Curve}
\newabbreviation{accuracy}{Acc}{Accuracy}
\newabbreviation{ate}{ATE}{Absolute Trajectory Error}

\newabbreviation{bev}{BEV}{Bird-Eye View}

\newabbreviation{clip}{CLIP}{Contrastive Language-Image Pretraining}
\newabbreviation{cnn}{CNN}{Convolutional Neural Network}

\newabbreviation{dof}{DoF}{DoF}

\newabbreviation{fov}{FoV}{Field of View}
\newabbreviation{fpr}{FPR}{False Positive Ratio}

\newabbreviation{gnn}{GNN}{Graph Neural Network}
\newabbreviation{gcn}{GCN}{Graph Convolutional Network}

\newabbreviation{knn}{KNN}{K-Nearest Neighbors}

\newabbreviation[plural=LLMs,firstplural=Large Language Models]{llm}{LLM}{Large Language Model}
\newabbreviation{lidar}{LiDAR}{Light Detection and Ranging}

\newabbreviation{mlp}{MLP}{Multi-Layer Perceptron}
\newabbreviation{mpc}{MPC}{Model Predictive Controller}
\newabbreviation{mse}{MSE}{Mean Squared Error}

\newabbreviation{ood}{OOD}{out-of-distribution}

\newabbreviation{pca}{PCA}{Principal Component Analysis}

\newabbreviation{rbf}{RBF}{Radial Basis Function}
\newabbreviation{rmp}{RMP}{Riemannian Motion Policies}
\newabbreviation{rmse}{RMSE}{Root Mean Square Error}
\newabbreviation{ros}{ROS}{Robot Operating System}
\newabbreviation{ros1}{ROS~1}{Robot Operating System}
\newabbreviation{roc}{ROC}{Receiver Operating Characteristic}
\newabbreviation{rpe}{RPE}{Relative Pose Error}
\newabbreviation{rf}{RF}{Random Forest}

\newabbreviation{sdf}{SDF}{Signed Distance Field}
\newabbreviation{slam}{SLAM}{Simultaneous Localization and Mapping}
\newabbreviation{sota}{SOTA}{state-of-the-art}
\newabbreviation{svm}{SVM}{Support Vector Machine}
\newabbreviation{svc}{SVC}{Support Vector Classifier}

\newabbreviation{vit}{ViT}{Vision Transformer}
\newabbreviation{vpr}{VPR}{Visual Place Recognition}
\newabbreviation{vlm}{VLM}{Vision-Language Model}

%% file: notation.tex
\newcommand{\pose}[3]{\mathbf{T}_{\mathtt{#1 #2}}_{#3}}
\newcommand{\rot}[3]{\mathbf{R}_{\mathtt{#1 #2}}_{#3}}
\newcommand{\pos}[2]{{\mathtt{_#1}} \mathbf{p}_{#2}}

\newcommand{\K}{\mathbf{K}_{3\times3}}

\newcommand{\img}[1]{\mathbf{I}^{#1}}

\newcommand{\feat}[1]{\mathbf{F}^{#1}}

\newcommand{\loss}[1]{\mathcal{L}_{\mathrm{#1}}}

\newcommand{\fun}[2]{f_{\mathrm{#1}}\left( #2 \right) }

\robustify{\pose}
\robustify{\rot}
\robustify{\pos}
\robustify{\K}
\robustify{\loss}
\robustify{\feat}
\robustify{\img}
\robustify{\fun}

\definecolor{TraversableBlue}{RGB}{49, 54, 149}
\definecolor{UntraversableRed}{RGB}{192, 26, 38}
\definecolor{PaperOrange}{RGB}{251, 151, 39}
\definecolor{PaperMagenta}{RGB}{150, 36, 145}
\definecolor{PaperBlue}{RGB}{67, 110, 176}
\definecolor{PaperCyan}{RGB}{66, 173, 187}

\usepackage{amssymb}

\usepackage{color}
\usepackage{xcolor}

%% file: table_FOV.tex
\begin{table}[tbp]
\setlength{\tabcolsep}{12pt}
\centering
\caption{Exp A - Translation and rotation RPE at 1m, averaged over five runs under different camera configurations.}
\label{tb:stat_RPE_FOV}
\begin{tabular}{cccc}
\toprule
\multicolumn{4}{c}{\textbf{Relative Pose Error (RPE) – Translation [m] / Rotation [°]}}                                                                                                                                                                                  \\ 
\midrule
\begin{tabular}[c]{@{}c@{}}FoV\\ (H $\times$ V)\end{tabular} & \begin{tabular}[c]{@{}c@{}}No.\\ Camera\end{tabular} & \begin{tabular}[c]{@{}c@{}}Translation \\ \gls{rmse} [m]\end{tabular} & \begin{tabular}[c]{@{}c@{}}Rotation\\ \gls{rmse} [°]\end{tabular} \\   
\midrule 
$64^{\circ}\times 90^{\circ}$       & 2                                              & Fail                                             & Fail                                            \\
$64^{\circ}\times 90^{\circ}$       & 3                                              & 0.61                                             & 3.57                                            \\
$92.4^{\circ}\times 126^{\circ}$         & 2                                              & 0.34                                            & \textbf{2.38}                                            \\
$92.4^{\circ}\times 126^{\circ}$         & 3                                             & \textbf{0.11}                                             & 2.70                      \\
\bottomrule                   
\end{tabular}
\end{table}

%% file: table_odom.tex
\begin{table}[tbp]
\centering
\caption{Exp B - Translation and rotation RPE (at \SI{1}{\meter} and \SI{5}{\meter}) for each odometry algorithm, averaged over five runs.}
\label{tb:stat_RPE_1m5m}
\resizebox{\columnwidth}{!}{
\begin{tabular}{clcccc} %
\toprule

 \multicolumn{6}{c}{\textbf{Relative Pose Error (RPE) – Translation [m] / Rotation [°]}} \\

\midrule

 \multirow{2}{0.5cm}{\vfill Dist}  &   \multicolumn{1}{c}{\multirow{2}{1cm}{\vfill\shortstack{Method}}}   &  \multicolumn{2}{c}{\emph{Seq. H2}} & \multicolumn{2}{c}{\emph{Seq. H3}}    \\

\cmidrule(lr){3-4}\cmidrule(lr){5-6}

&  & \multicolumn{1}{c}{\shortstack{Translation \\ \gls{rmse} }} & \multicolumn{1}{c}{\shortstack{Rotation \\ \gls{rmse} }}  &  \multicolumn{1}{c}{\shortstack{Translation \\ \gls{rmse}}} & \multicolumn{1}{c}{\shortstack{Rotation \\ \gls{rmse} }} \\

\cmidrule(lr){1-6}

 \multirow{3}{*}{\shortstack{1m}}&\multirow{1}{1.5cm}{ORB-SLAM}   &  \multicolumn{1}{c}{0.08 } & \multicolumn{1}{c}{\textbf{3.14 }}& \multicolumn{1}{c}{0.19}   & \multicolumn{1}{c}{\textbf{2.49} } \\ 

& \multirow{1}{*}{OpenVINS}    &  \multicolumn{1}{c}{\textbf{0.06}} &
\multicolumn{1}{c}{3.45} & \multicolumn{1}{c}{0.15}  & \multicolumn{1}{c}{2.55
} \\ 

& \multirow{1}{*}{VILENS-MC}    &  \multicolumn{1}{c}{0.10} &
\multicolumn{1}{c}{3.62} & \multicolumn{1}{c}{\textbf{0.14}} &
\multicolumn{1}{c}{2.55}
\\

\midrule

 \multirow{3}{*}{\shortstack{5m}} & \multirow{1}{1.5cm}{ORB-SLAM}   &
 \multicolumn{1}{c}{0.27} & \multicolumn{1}{c}{3.62}&
 \multicolumn{1}{c}{0.58}   & \multicolumn{1}{c}{\textbf{3.56}} \\ 

& \multirow{1}{*}{OpenVINS}    &  \multicolumn{1}{c}{\textbf{0.20}} &
\multicolumn{1}{c}{3.40} & \multicolumn{1}{c}{0.40} & \multicolumn{1}{c}{4.10}
\\ 

& \multirow{1}{*}{VILENS-MC}   &  \multicolumn{1}{c}{0.26} &
\multicolumn{1}{c}{\textbf{3.20}} & \multicolumn{1}{c}{\textbf{0.38}}  &
\multicolumn{1}{c}{4.25} \\
\bottomrule
\end{tabular}}
\end{table}

%% file: table_accuracy.tex
\begin{table}[tbp]
\setlength{\tabcolsep}{3pt}
\centering
\caption{Exp D - Terrain Reconstruction Quality. We measured the point-to-point distance between our terrain reconstruction and the ground truth point cloud scan. We present the mean, max and $90^{\text{th}}$ percentile error to quantify the mapping quality.}
\label{tb:accuracy}
\resizebox{0.85\columnwidth}{!}{%
\begin{tabular}{cccccc}
\toprule
\multicolumn{5}{c}{\textbf{Point-to-point Distance of Staircase Reconstruction}} \\
\midrule
dataset  & res. [cm] & mean [cm] 
 & max [cm]
 & 90\% [cm]\\
\midrule
\multirow{1}{0.5cm}{\textit{Human}}  & 2 & 1.36 & 8.45 & 2.84\\
\midrule
\multirow{1}{*}{\shortstack{\textit{Exo}}}  & 2 & \multicolumn{1}{c}{0.80} & \multicolumn{1}{c}{6.64} & \multicolumn{1}{c}{1.85} \\
\bottomrule
\end{tabular}
} 
\end{table}